\documentclass[sigconf,natbib=true,screen,authorversion,nonacm]{acmart}
\usepackage{fancyhdr}
\AtBeginDocument{%
    \addtolength{\footskip}{2.0\baselineskip}%
    \fancyfoot[L]{© Authors 2022. This is the author's version of the work. It is posted here for your personal use. Not for redistribution. The definitive version was published in SIGIR 2022, https://doi.org/10.1145/3477495.3531858}%
}

\AtBeginDocument{%
  \providecommand\BibTeX{{%
    \normalfont B\kern-0.5em{\scshape i\kern-0.25em b}\kern-0.8em\TeX}}}

\copyrightyear{2022}
\acmYear{2022}
\setcopyright{acmlicensed}
\acmConference[SIGIR '22]{Proceedings of the 45th International ACM SIGIR Conference on Research and Development in Information Retrieval}{July 11--15, 2022}{Madrid, Spain.}
\acmBooktitle{Proceedings of the 45th International ACM SIGIR Conference on Research and Development in Information Retrieval (SIGIR '22), July 11--15, 2022, Madrid, Spain}
\acmPrice{15.00}
\acmISBN{978-1-4503-8732-3/22/07}
\acmDOI{10.1145/3477495.3531858}

\begin{document}
% \fancyhead{}
\title{Which Discriminator for Cooperative Text Generation?}

\author{Antoine Chaffin}
\affiliation{%
  \institution{IRISA, IMATAG}
  \city{Rennes}
  \country{France}
}
\email{antoine.chaffin@irisa.fr}
\email{antoine.chaffin@imatag.com}

\author{Thomas Scialom}
\affiliation{%
  \institution{ISIR - Sorbonne Université, reciTAL}
  \city{Paris}
  \country{France}
}
\email{thomas@recital.ai}

\author{Sylvain Lamprier}
\affiliation{%
  \institution{ISIR - Sorbonne Université}
  \city{Paris}
  \country{France}
}
\email{sylvain.lamprier@isir.upmc.fr}

\author{Jacopo Staiano}
\affiliation{%
  \institution{reciTAL}
  \city{Paris}
  \country{France}
}
\email{jacopo@recital.ai}

\author{Benjamin Piwowarski}
\affiliation{%
  \institution{CNRS, ISIR - Sorbonne Université}
  \city{Paris}
  \country{France}
}
\email{benjamin.piwowarski@lip6.fr}

\author{Ewa Kijak}
\affiliation{%
  \institution{Université Rennes, IRISA}
  \city{Rennes}
  \country{France}
}
\email{ewa.kijak@irisa.fr}

\author{Vincent Claveau}
\affiliation{%
  \institution{CNRS, IRISA}
  \city{Rennes}
  \country{France}
}
\email{vincent.claveau@irisa.fr}

\renewcommand{\shortauthors}{Chaffin et al.}

\begin{abstract}
Language models generate texts by successively predicting probability distributions for next tokens given past ones. 
A growing field of interest tries to leverage external information in the decoding process so that the generated texts have desired properties, such as being more natural, non toxic, faithful, or having a specific writing style. 
A solution is to use a classifier at each generation step, resulting in a cooperative environment where the classifier guides the decoding of the language model distribution towards relevant texts for the task at hand. 
In this paper, we examine three families of (transformer-based) discriminators for this specific task of cooperative decoding: bidirectional, left-to-right and generative ones. We evaluate the pros and cons of these different types of discriminators for cooperative generation, exploring respective accuracy on classification tasks along with their impact on the resulting sample quality and computational performances. 
We also provide the code of a batched implementation of the powerful cooperative decoding strategy used for our experiments, the Monte Carlo Tree Search, working with each discriminator for Natural Language Generation.
\end{abstract}



% %%
% %% Keywords. The author(s) should pick words that accurately describe
% %% the work being presented. Separate the keywords with commas.
% \keywords{datasets, neural networks, gaze detection, text tagging}

\sloppy

\maketitle
\section{Introduction}

Transformer~\cite{DBLP:conf/nips/VaswaniSPUJGKP17} architectures, coupled with an increase in computing capabilities,  allows current Language Models (LM) to generate very plausible texts. Given an initial sequence of tokens (the prompt), the LM computes a probability distribution for the next token. A token is then sampled from this distribution and added to the initial sequence to generate the following token auto-regressively. Choosing the next token given the distribution (decoding) is commonly done using greedy search, beam search~\cite{dept._2018} or top-k/p sampling~\cite{DBLP:conf/acl/LewisDF18, DBLP:conf/iclr/HoltzmanBDFC20}; they select the next token only based on the likelihood (according to the LM) of the resulting sequence, which offers only limited control over the text finally generated. % at the end of the decoding  process. %limiting the control over the generated text to  the   original prompt.

Yet, large LMs trained with non curated data are known to produce toxic and inappropriate content~\cite{DBLP:conf/fat/BenderGMS21, DBLP:conf/emnlp/GehmanGSCS20}. This is particularly problematic for Information Retrieval tasks that imply text generation, such as question-answering from the Web \cite{DBLP:journals/corr/abs-2112-09332,NEURIPS2020_6b493230}, query-focused  multi-documents summarization \cite{pasunuru2021data}, query expansion \cite{ClaveauWI2021}, query suggestion \cite{DBLP:journals/tois/MustarLP22}, or chatbots for interactive search \cite{pallagani2021generic}, which leverage contents from various -- and sometimes untrusted -- information sources.   %[a compléter avec les besoins pour la RI etc]

%Discriminators, corresponding to trained classifiers,
Classifiers can be trained to identify a specific property of a text and thus provide useful information to guide the LM towards the desired property. 
For instance, following Generative Adversarial Networks~\cite{DBLP:journals/cacm/GoodfellowPMXWO20}, many studies train binary  discriminators to distinguish real from generated contents, to approximate distributions of observed documents \cite{yu2017seqgan}. Other studies train classifiers on semantic properties such as polarity to learn the generation process towards positive or negative texts~\cite{DBLP:journals/corr/abs-2109-13582, DBLP:conf/emnlp/KrauseGMKJSR21, DBLP:conf/iclr/DathathriMLHFMY20}. 
In the context of Information Retrieval, this might also be used for instance to increase relevance of synthetic answers w.r.t. to the user's query. 
For all these purposes, there is an increasing interest for discriminator-generator cooperative decoding, where discriminators are used to guide generation %the decoding at generation time %of text sequences
\cite{DBLP:conf/acl/ChoiBGHBF18, DBLP:conf/icml/ScialomDLPS20, DBLP:conf/eacl/GabrielBDHBLCC21, DBLP:conf/iclr/DengBOSR20}.          
Currently, top performing discriminators are transformers using bidirectional attention~\cite{DBLP:conf/naacl/DevlinCLT19}, but this does not fit the iterative nature of the generation process. Indeed, it requires to recompute every hidden state of the whole sequence for any additional token, preventing the use of cached hidden states and resulting in a quadratic cost w.r.t. the sequence length at each timestep. 
On the other hand, unidirectional transformers, which employ left-to-right masks to only depend on past tokens for text  encoding/decoding~\cite{Radford2019LanguageMA}, induce hidden states that can be reused for subsequent steps, hence involving linear computing complexity. However, these two types of discriminators only score one sequence at a time, given as input of the model.  
This limits the number of possible tokens to be considered at each decoding step, to avoid a computationally prohibitive cost. Solving this issue, recently introduced Generative Discriminators  (GeDi)~\cite{DBLP:conf/emnlp/KrauseGMKJSR21} give scores for all tokens from the vocabulary at once, hence dramatically reducing the cost of width exploration. In this paper, we explore the pros and cons of these three types of discriminators (bidirectional, unidirectional, generative) when used in cooperative language decoding.

% about the MCTS framework
In parallel, approaches relying on Monte Carlo Tree Search algorithm (MCTS)~\cite{DBLP:conf/cg/Coulom06}  
have been used for cooperative generation with more sophisticated exploration strategies than beam search.
% Recently, using the Monte Carlo Tree Search algorithm (MCTS)~\cite{DBLP:conf/cg/Coulom06} to define non-myopic discriminator-guided decoding, various studies demonstrate % to sample the next token in order to maximize the discriminator score led to 
This non-myopic discriminator-guided decoding lead to
state-of-the-art results in different applications~\cite{selfGAN, DBLP:journals/corr/abs-2109-13582, DBLP:conf/emnlp/LeblondASPLASV21, DBLP:journals/corr/abs-2201-12320}. We therefore use this promising cooperative decoding approach for our experiments and provide an implementation of the MCTS that allows to generate texts in batch for each type of discriminator\footnote{\href{https://github.com/NohTow/PPL-MCTS/tree/main/teammates}{https://github.com/NohTow/PPL-MCTS/tree/main/teammates}} based on the HuggingFace's transformers library~\cite{DBLP:conf/emnlp/WolfDSCDMCRLFDS20}. 
% For our experiments, we adopt the recently proposed Monte Carlo Tree Search algorithm as the framework to test the discriminators for decoding. 
% MCTS provides non-myopic discriminator-guided decoding, yielding
% state-of-the-art results in different applications~\cite{selfGAN, DBLP:journals/corr/abs-2109-13582, DBLP:conf/emnlp/LeblondASPLASV21, DBLP:journals/corr/abs-2201-12320}.
%

%about our results
\begin{comment}
Results show that, while GeDi would allow better performance with wide tree search structures, with even a better robustness regarding out of domain texts, the intrinsic depth-first search process of MCTS provides a clear advantage to standard discriminators, especially the greatly more practicable unidirectional one.  
\end{comment}
Before exposing our experimental study, we further define the task of cooperative decoding and justify our study in the next section.
%%%%%%%%%%%%%%%%%%%%%%%%%%%%%%%%%%%%%%%%%%%%%%%%%%%%%%%%%%%%%%%%%%%%%%%%%%%%%%
\section{Background and Motivations}
\label{sec:background}

\subsection{Cooperative Decoding with MCTS}
% In cooperative text generation, information from the discriminator is combined to the generator distribution to skew the generation towards the desired property defined by one class of the discriminator. 
% Inspired from Value-guided beam search  \cite{he2017decoding,Ren_2017_CVPR}, but using class discriminators rather than value networks,  Discriminative Adversarial Search (DAS)~\cite{DBLP:conf/icml/ScialomDLPS20} proposed to  %this can be done by re-ranking the 
% re-rank beam-generated sequences according to their discrimination scores. % likelihood values augmented a based on its scores as in Discriminative Adversarial Search (DAS)~\cite{DBLP:conf/icml/ScialomDLPS20}, where the likelihood in the beam search is augmented by the score of the discriminator.
% This approach paved the way for many other studies on cooperative decoding \cite{chen2020adding,DBLP:conf/iclr/DengBOSR20,yuan2021event}. %  \cite{DBLP:conf/iclr/DengBOSR20, }, 
In cooperative text generation, information from the discriminator is combined to the generator distribution to skew the generation towards the desired property defined by one class of the discriminator~\cite{chen2020adding,DBLP:conf/iclr/DengBOSR20,yuan2021event, DBLP:conf/acl/ChoiBGHBF18, DBLP:conf/icml/ScialomDLPS20}. For instance, inspired from Value-guided beam search  \cite{he2017decoding,Ren_2017_CVPR}, but using class discriminators rather than value networks,  Discriminative Adversarial Search (DAS)~\cite{DBLP:conf/icml/ScialomDLPS20} proposed to  %this can be done by re-ranking the 
re-rank beam-generated sequences according to their discrimination scores. % likelihood values augmented a based on its scores as in Discriminative Adversarial Search (DAS)~\cite{DBLP:conf/icml/ScialomDLPS20}, where the likelihood in the beam search is augmented by the score of the discriminator.
% This approach paved the way for many other studies on cooperative decoding \cite{chen2020adding,DBLP:conf/iclr/DengBOSR20,yuan2021event}. %  \cite{DBLP:conf/iclr/DengBOSR20, }, 
Among these approaches, MCTS-based ones~\cite{selfGAN, DBLP:journals/corr/abs-2109-13582, DBLP:conf/emnlp/LeblondASPLASV21, DBLP:journals/corr/abs-2201-12320} allowed to obtain state-of-the-art results in various NLG tasks, by overcoming the limitations of myopic %propose to overcome the curse of 
left-to-right decoding (and difficult value-network learning~\cite{DBLP:conf/emnlp/LeblondASPLASV21}). %µ, 
%The major drawback of such approach is the lack of long-term vision which results in sequences which turns out to be bad after a few generation steps.

% MCTS is an algorithm that will iteratively build a tree in order to take short-term decision that are promising in the long run. During each iteration, a search toward an unexplored node is driven by a compromise between exploiting good sequences and exploring promising ones which can be defined by the hyperparameter $c_{puct}$. 
% The corresponding sequence is then scored by the discriminator and this score is back-propagated to every of its sub-sequences.
% Hence, next token scores are based on scores of possible continuations, allowing to get a better overview of the what is achievable for a given choice.
MCTS \cite{silver2017mastering} is an algorithm that iteratively builds a (generation) tree at each decision  step, 
to take short-term decisions that might be promising in the long run. Each iteration is composed of three steps. First, during \textbf{selection}, a search toward an unexplored node is driven by a compromise between exploiting good partially generated sequences and exploring promising ones. This trade-off %. This compromise
is controlled %defined 
by the parameter $c_{puct}\in\mathbb R$ (higher values mean more exploration). Then,  \textbf{expansion} is  performed by creating children of the selected node. Finally, the corresponding sequence is scored by the discriminator and the score of every parent up to the tree root are updated accordingly during a \textbf{backpropagation} phase. 
%Hence, next token scores are based on scores of possible continuations, allowing to get a better overview of the what is achievable for a given choice.
%Note that the discriminator can be used after an additional \textbf{roll-out} step, where the sequence to score is first filled with additional tokens, further enhancing the long-term vision.
In MCTS, this back-propagated score is usually computed from the selected node by \textbf{rolling out} until a terminal node and by evaluating the resulting full sequence. As done in other approaches for cooperative decoding, we replace these costly roll-outs in our experiments by scores provided by discriminators trained on unfinished text sequences. %\cite{selfGAN}. 
In this work, we experiment on which kinds of discriminators are the best cooperative partners for generating with MCTS. %(mostly with MCTS, but we also discuss their impact on other common generation frameworks).

\subsection{Choosing the Right Teammate}

By default, attention layers as defined in~\cite{DBLP:conf/nips/VaswaniSPUJGKP17} are bidirectional: every token can attend to tokens at every position. 
When it comes to discrimination, models based on such bidirectional attention are commonly used since "intuitively, it is reasonable to believe that a deep bidirectional model is strictly more powerful than [...] a left-to-right model"~\cite{DBLP:conf/naacl/DevlinCLT19}.
However, while it %bidirectionality
brings some capacity to the model, it also makes  it non auto-regressive: when a token is added at the end of a sequence, every hidden states need to be re-computed. 

One way to train a transformer based LM for text generation is to use unidirectional attention masks \cite{Radford2019LanguageMA}. In this unidirectional setting, any extra token added at the end of a sequence does not change the already calculated hidden states, since previous tokens do not attend to it. %this new token.
Thus, starting %at step $t$ 
from an already classified sequence $x_{1:t-1}$, classifying $x_{1:t}$ %
%
%
%if the sequence of length $t-1$, $x_{1:t-1}$ has been classified, doing the inference on the same sequence with a new token added, $x_{1:t}$
only requires to compute $t$ attention scores, rather than the whole set of $t^2$ scores per self-attention layer, as it would be required in the bidirectional setting.
In common discriminative tasks, this does not matter since only entire sequences are discriminated. Hence, none of the hidden states needs to be reused for another next sample. 
However, for a use in auto-regressive cooperative decoding,  %in auto-regressive generative tasks,
where input sequences are often the continuation of already discriminated ones % as in DAS sequence,  %which is the case in our cooperative decoding setting, 
unidirectional attention allows to reuse contextual encoding of previous tokens, %enables to limit computation %only requires to compute attention on the token added at the previous step, 
hence greatly speeding up the process.

However, even with unidirectional discriminators, evaluating every %beam's
possible continuation of a given sequence is intractable since, for a vocabulary of size $|\mathcal{V}|$, it requires $|\mathcal{V}|$ forward passes at each decoding step. %each beam.
$|\mathcal{V}|$ being in the order of ten thousand, discriminating every possible continuation of decoding sequences is too costly. %of every beam at every time step
% is just way too costly, even with the linear complexity of unidirectional models. 
% Thus, cooperative approaches such as DAS circumvent this issue by pre-filtering %do not perform an exact beam search and
% %first filter
% potential continuations on the nucleus of the %only tokens with the highest likelihood for the 
% language model distribution \cite{nucleusSampling}. This choice necessarily biases the resulting generated distribution. %search towards the likelihood.
Thus, cooperative approaches have to circumvent this issue by limiting the number of sequences actually evaluated by the discriminator. For example, DAS pre-filters potential continuations on the nucleus of the LM distribution \cite{DBLP:conf/iclr/HoltzmanBDFC20}. This choice necessarily biases the resulting generated distribution.

Recently, \cite{DBLP:conf/emnlp/KrauseGMKJSR21} introduced Generative Discriminators (GeDi) that exploit Class-Conditionnal Language Models (CC-LMs)~\cite{DBLP:journals/corr/abs-1909-05858} to discriminate every token at once. CC-LMs %, instead of only conditioning %modeling
%the probability of the next token at timestep $t$ given the $t-1$ previous ones $p(x_{t} \mid x_{1:t-1})$,
condition %this
distributions of sequence $x$ on a desired class  of interest $c$:  $p(x \mid c) = \prod_t p(x_{t} \mid x_{1:t-1}, c)$. %In this setting, applying the chain rule gives $\prod_{t=1}^{T} p(x_{t} \mid x_{1:t-1}, c)=p(x_{1:T} \mid c)$. 
%Given a set of class $\mathcal{C}$ that forms a partition, 
Assuming a uniform prior distribution of classes $c \in {\mathcal C}$, Bayes' rule enables %gives a relation that allows
to use this for discrimination: $p(c \mid x_{1:T}) \propto p(x_{1:T} \mid c)$.  %  p(c)$. %}{p(x_{1:T})}$.
%Let us assume a set of class $\mathcal{C}$ that forms a partition, the law of total probability gives ${p(x_{1:T})} = \sum_{c^{\prime} \in \mathcal{C}} p(c^{\prime})p(x_{1:T} \mid c^{\prime})$. CC-LMs can be used as discriminators: $p(c \mid x_{1:T})=\frac{p(x_{1:T} \mid c) p(c)}{\sum_{c^{\prime} \in C} p(c^{\prime})p(x_{1:T} \mid c^{\prime})}$. 
%Assuming an uniform distribution of classes, %i.e $p(c)=p(c^{\prime}) \forall c, c^{\prime} \in \mathcal{C}$, it simplifies to $\frac{p(x_{1:T} \mid c)}{\sum_{c^{\prime} \in C} p(x_{1:T} \mid c^{\prime})}$. Since CC-LMs gives the distribution of $p(x_{t} \mid x_{1:t-1}, c)$ over the vocabulary, %it comes down to simply define the posterior probability class  distributions 
Thus, it only requires $|\mathcal{C}|$ forward passes % conditioning on each class 
to get the discrimination scores of all possible sequence continuations. % possible. 
%In the DAS example, 
$|\mathcal{C}|$ being usually much %hundred times
lower than $|\mathcal{V}|$, this makes the consideration of every token tractable for sequential discriminative decoding. To improve discriminatory capacity of such models, training of CC-LMs used in GeDi leverages a discriminative loss $\mathcal{L}_{d}$ in addition to the traditional language modeling loss $\mathcal{L}_{g}$. This discriminative loss corresponds to a cross-entropy loss using the model as a discriminator and a  hyper-parameter $\lambda$ is used to define the balance between the two objectives: $\mathcal{L}_{total}=\lambda \mathcal{L}_{g}+(1-\lambda) \mathcal{L}_{d}$.

These three types of discriminators offer different capacity / complexity trade-off, which are studied in this paper for cooperative decoding with MCTS. 
More precisely, three questions are explored: 1) 
%\begin{itemize}
    %\item
    How these models differ in pure discrimination accuracy?  
    %\item
    2) To which extend are these differences noticeable in generated texts? %\item
    3) How do these methods compare in terms of computation complexity for cooperative decoding? %What is the empirical computation gains for state-of-the-art cooperative generation method ?
%\end{itemize}

% Explication plus rigoureuse du gain via l'attention unidirectionnel et GeDi
% Analyse sur une méthode de re-ranking simple (type DAS) où gain théorique == gain empirique 

%%%%%%%%%%%%%%%%%%%%%%%%%%%%%%%%%%%%%%%%%%%%%%%%%%%%%%%%%%%%%%%%%
\section{Empirical Study} %xperiments}

According to previous studies, unidirectional models should yield worse accuracy than bidirectional ones~\cite{DBLP:conf/naacl/DevlinCLT19, Radford2019LanguageMA} and better than discriminative generators~\cite{DBLP:journals/corr/YogatamaDLB17, DBLP:conf/nips/NgJ01}.
To thoroughly assess the pros and cons of these models using state-of-the-art transformer  architectures, it is crucial that the only difference is the studied property (uni- vs bi- directionality, and discriminative vs generative). Thus, we propose to use the same %BERT 
backbone for all settings to prevent %any undesirable biases due to
any external  confounding factors, with a single fully connected output layer 
on top of the  contextual embedding of the last token 
to produce discrimination scores.  
Starting from BERT~\cite{DBLP:conf/naacl/DevlinCLT19} as bidirectional discriminator, a triangular self-attention mask is applied for adapting it from the bidirectional to the unidirectional setting in our experiments, following~\cite{DBLP:conf/nips/00040WWLWGZH19}.
%Following~\cite{DBLP:conf/nips/00040WWLWGZH19},  a triangular self-attention mask is applied to BERT for adapting it from the bidirectional to the unidirectional setting in our experiments. %to make it left-to-right for . %This is the only difference between unidirectional and bidirectional models used in our experiments.
%Discrimination scores are provided in every setting by %Only a 
%a fully connected layer of shape  $(hidden\_size, num\_classes)$ on top of the BERT contextual  embedding of the last token of the sequence. %is added to make the discrimination based on BERT hidden state. 
Then, the generative discriminator is the same as the left-to-right one, the only difference being the size %the size
of the output layer that changes from  %shape
$(hidden\_size, num\_classes)$ to  %becomes  
$(hidden\_size, vocab\_size)$. %Using such a same BERT backbone and a single fully connected output layer prevents any confounding factor, allowing to fairly study  the gain and losses of each type of model.

%A common practice to use BERT for discrimination is to use the hidden state of the [CLS] token added to the beginning of the sequence, however, in the left-to-right case, this hidden state is not updated with the context of the sequence. To make the processing of each network as close as possible, hidden state of the last token is used as input of the last layer, in accordance with common practices in unilateral models. Using the average of every token hidden states is also a possibility, yet we did not observe a difference in preliminary experiments.

Experiments are made on two datasets from~\citep{DBLP:conf/nips/ZhangZL15}: \href{https://huggingface.co/datasets/amazon_polarity}{amazon\_polarity} which is a binary (positive or negative) online reviews classification task and \href{https://huggingface.co/datasets/ag_news}{AG\_news} which is a topic classification task with 4 labels (\texttt{world}, \texttt{sport}, \texttt{business} and \texttt{science}). These datasets allows to study results of cooperative generation on two rather different constraints and domains: applying polarity on online reviews and writing news about a specific topic. Also, AG\_news allows to study the generalization to non-binary classification and texts with more diverse content.
%Using the same BERT backbone and a single fully connected output layer prevents any confounding factor, allowing to study precisely the gain and losses of each type of model.
%For the same purpose, the training is also standardized. 
Each model is trained for 20 epochs using AdamW~\cite{DBLP:conf/iclr/LoshchilovH19} with HuggingFace's trainer default parameters ($\beta_1 = 0.9, \beta_2= 0.999, eps=1e-8$) and a linear scheduler with no warmup. %The final discriminator is the one achieving the best validation accuracy. 
Batch size is set to the maximum that can fit in the memory of a Quadro RTX 6000 during GeDi training (4 for AG\_news and 8 for amazon\_polarity). Gradient accumulation is set to emulate a batch size of 128. For training GeDi, we set $\lambda = 0.6$ according to the authors (and did not observe significant difference when setting $\lambda = 0$ to strengthen the classification capacity).

% Ajouter des informations sur le training (dire que c'est les même opti/schedu et donner les paramètres)

%En sachant ce qu'on peut gagner en coût de calcul, la question qui se pose est le prix à payer pour ce gain (no free lunch) -> Etude de l'accuracy des différents classifiers, notamment en fonction de la taille de l'input, caractéristique importante pour la génération coopérative

\begin{figure*}[tb!]
\includegraphics[width=0.8\textwidth]{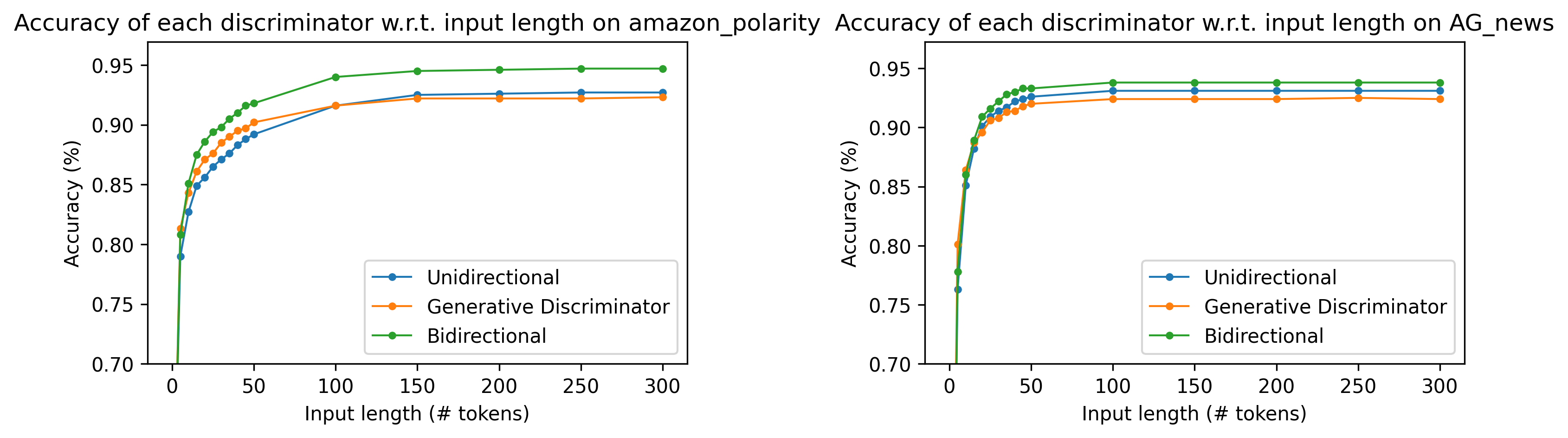}
\caption{Accuracy (\%) of the different type of discriminators w.r.t. the input length (\# tokens)}
\label{fig:raw_accuracy}
\end{figure*}

\subsection{Discrimination Strength}

For discriminators, accuracy has an utmost importance: it defines how well it solves the intended task. In the context of cooperative generation, having a good accuracy on complete sequences is not sufficient: an informative output with uncomplete sequences is needed so that the discriminator can be used throughout the generation process. %For example, in the case of DAS \cite{DBLP:conf/icml/ScialomDLPS20}, beams incorrectly scored in the beginning will cause the generation process to deviate from optimal solutions. 
Thus, plotting the accuracy w.r.t. the number of input tokens gives information about the capacity of the model to guide the generation at different timesteps, the main property expected for discriminators in cooperative generation. 
Note that, following common practice in cooperative generation, the discriminator is trained on sequences of variable lengths to avoid a mismatch between training and test tasks. %, allowing to use the discriminator on unfinished sequences during the generation process.
%It should be noted that, a common practice in cooperative generation is to train the discriminator on sequences truncated at random lengths to avoid a mismatch between training and test tasks. We follow this practice for our experiments and study the relevance of such training in a next sub-section.

Results reported in Fig.~\ref{fig:raw_accuracy} show that every discriminator exhibits the same behavior: starting from random predictions, accuracy quickly increases with the input length until reaching a plateau. %corresponding to the accuracy usually reported in standard classification tasks. 
The expected ordering is observed: bidirectional models perform better than unidirectional models, which themselves perform better than generative ones. However, it should be noted that the gap is rather small and only appears when approaching the plateau. Favoring bidirectional models in accuracy-critical tasks is justified, but it is not necessarily clear that these small differences will reflect in the quality of cooperatively generated texts. %Note at last that complementary   experiments, on random sequences to be discriminated, showed however that GeDi is more robust to out of domain sequences, with discrimination scores greatly closer to maximal uncertainty (i.e., $D(x)=0.5$),  than with discriminative models, which tend to greatly favor one class over the other ones in such cases. However, this may not impact results in generation, since such random samples may be ignored during MCTS decoding,  %not be encountered in learning, 
%depending on its exploration/exploitation trade-off. % of MCTS. % avoided from can not be Experiments in next sections will inspect if that  

Please note that this corresponds to the accuracy on \textit{in domain} data, and that complementary - non reported - experiments, on random sequences to be discriminated, showed however that GeDi is more robust to \textit{out of domain} sequences:  its discrimination scores are greatly closer to maximal uncertainty (i.e., $p(c|x)=0.5$),  than those of discriminative models  which tend to greatly favor one class over the other ones in such cases. However, this may not impact results in generation, since such random samples are not likely to be observed during MCTS decoding, because of the language model prior guiding search towards in distribution sequences.
% Explication sur l'accuracy (de GeDi) : on évalue seulement la classification pour UNE sequence, donc là en soit on test pas vraiment la propriété de "score tout le vocabulaire". Ceci dit, la classification est faite à partir de ces probas, donc ça montre quand même que, pour les séquences "normales", l'accuracy est pas mauvaise. On a pas d'information sur les probas en OOD, mais on peut supposer que si on se guide aussi via le likelihood, on devrait rester sur une partie de l'espace qui est bien géré. Quid sinon ?

\subsection{Generation Quality}
%Results show that there is a relatively small difference in the classification accuracy of these models. However, its impact on the text generated is still unknown. 
%The impact of relatively small differences in classification accuracies of the different type of discriminators is unknown. The additional misclassifications might add up and cause the generation process to deviate a lot or the difference might be small enough to not be noticeable when used for generation.
%To study the effective difference in final samples, each discriminator is tested in a Monte Carlo Tree Search (MCTS) decoding setup. MCTS has been recently shown to be really effective for cooperative generation~\cite{selfGAN, DBLP:journals/corr/abs-2109-13582, DBLP:conf/emnlp/LeblondASPLASV21}, allowing to achieve state-of-the-art results in considered tasks. 
To assess whether the -- relatively small --  differences in classification accuracies impact the results on cooperative generation with MCTS, 
we follow the PPL-MCTS~\cite{DBLP:journals/corr/abs-2109-13582} setup by constraining the generation process towards a desired class $c$ using $p(c|x)$ given by the considered discriminator. 
%To study the quality of the guiding signal brought by the discriminator, we report the oracle (a model trained on disjoint data) accuracy, i.e the percentage of generated texts belonging to the target class according to the oracle, as well as the oracle perplexity to control the written quality and Self-BLEU~\cite{DBLP:conf/sigir/ZhuLZGZWY18} values to ensure enough diversity across samples. 
Automatic metrics are used to study the quality of the guiding signal brought by the discriminator: 1) Accuracy corresponds to the average rate of generated sequences for any class $c$ to be correctly classified as $c$ by an % confor any to be  classificof an
oracle %(a model trained on disjoint data) 
discriminator trained on disjoint data, 2) Self-BLEU~\cite{DBLP:conf/sigir/ZhuLZGZWY18} focuses on % ensure enough 
diversity across samples, by measuring BLEU scores between %pairs of
generated sequences, and 3) Oracle perplexity stands for the perplexity of an oracle LM trained on disjoint data, allowing to control the writing quality of generated texts.   % to assert if generated texts belong to the target class. 
We used a bidirectional BERT model as oracle discriminator to get the most accurate evaluation possible. Language models are also BERT models with an LM head in order to use the same tokenizer. 
%For both oracles, we used a bidirectional BERT model to get the most accurate evaluation as possible. %A language model oracle is also trained to control the written quality by reporting perplexity. Both the language model used for generation and the oracle are BERT model with LM heads. Finally, . 
Average results over 500 sampled test texts using each type of discriminator on the two datasets are reported in Tab.~\ref{tab:generation_quality}. We also report results obtained using the %perplexity of the
vanilla LM likelihood $p(x)$ as back-propagated score in MCTS evaluation, to provide baseline 
results achievable without discriminators. Results are obtained using best performing hyper-parameters in the literature ($c_{puct}=3$, temperature $\tau=1$) and 50 iterations of MCTS per token, unless specified otherwise. We report statistical significance between each type of discriminator using t-test with p-value=0.01.

The difference of %oracle
generation accuracy when using bidirectional and unidirectional discriminator shows that the difference in raw accuracy reflects in resulting samples when used for cooperative generation. The higher difference on amazon\_polarity also results in a higher difference in %oracle
generation accuracy. However, this difference is relatively limited and the generation does not seem to deviate too much using unidirectional discriminators. 
Results using generative discriminators are different, with a % The 
significantly greater drop of accuracy %is significantly higher 
than between uni- and bi-directional models on AG\_news,  although %whereas
the gap in raw accuracy is similar. More surprising is the result on amazon\_polarity where, despite  similar raw accuracies, we observe a 10-point drop % results in a difference
of generation accuracy. We hypothesize that this is because the signal is not as informative: while raw accuracies are pretty similar, the average score attributed to the ground truth class in evaluation is significantly lower in the case of GeDi. This means that its signal promotes less good solutions %and more bad ones
than standard discriminators when guiding the generation. 
The type of discriminator %does not really have a
has no significant impact on the other metrics. Please note that the general difference of Self-BLEU and oracle perplexity between the two datasets is due to the difference in their content: AG\_news is more diverse, which results in lower Self-BLEU and higher perplexity. % reported on AG\_news. On amazon\_polarity, we observe that 
Finally, we notice that doubling the number of MCTS iterations %to $100$ 
allows to increase the accuracy results of the unidirectional model, %reaching the performances of the bidirectional one 
bridging the gap between both model for a still lower computational cost (see next section).  
%It should be noted that MCTS allows to set a compromise between compute and quality. The speed-up brought by, for example, the unidirectionality, can be reinvested in additionnal iteration of the algorithm.

%It should be noted that MCTS allows to set a compromise between compute and quality. The speed-up brought by, for example, the unidirectionality, can be reinvested in additionnal iteration of the algorithm, bridging the gap between both model as shown in Table~\ref{tab:generation_quality}.

\begin{table*}[htb]
\resizebox{0.9\textwidth}{!}{
\begin{tabular}{llcccccccc}
                                         & \multicolumn{3}{c}{amazon\_polarity}                                                                                   & \multicolumn{3}{c}{AG\_news} \\
\multicolumn{1}{l|}{Value}          & \multicolumn{1}{c}{Accuracy ↑}        & 5 - Self-BLEU  ↓                      & \multicolumn{1}{c|}{Oracle perplexity ↓}                         & Accuracy ↑                 & 5 - Self-BLEU  ↓                       & \multicolumn{1}{c}{Oracle perplexity ↓}                 \\ \hline
\multicolumn{1}{l|}{$p(x)$}   & 70.8                            & 0.652                           & \multicolumn{1}{c|}{10.49} & 86.6 & 0.306                 & 29.08  \\ \hline

\multicolumn{1}{l|}{Bidirectional}   & \textbf{$96.0^\ast$}                            & $0.531^\ast$                           & \multicolumn{1}{c|}{12.25} & \textbf{$94.8^\ast$} & 0.319                 & 29.13 \\
\multicolumn{1}{l|}{Unidirectional}   & $93.0^\ast$                            & $0.528^\ast$                           & \multicolumn{1}{c|}{11.98} & 93.4 & 0.313                 & 29.99  \\
\multicolumn{1}{l|}{Unidirectional (100 its)}   & $93.6^\ast$                            & \textbf{$0.522^\ast$}                           & \multicolumn{1}{c|}{10.73} & $94.6^\ast$ & 0.323 & 30.92 \\
\multicolumn{1}{l|}{Generative discriminator}   & 84.4                            & 0.576                           & \multicolumn{1}{c|}{11.92} & 91.8 & 0.321                & 29.43  \\

\end{tabular}
}
\caption{Performance of MCTS w.r.t. the metric to optimize on amazon\_polarity (left) and AG\_news (right) datasets. $\ast$ indicates statistically significant improvement against Generative Discriminator. Note that no model demonstrated significant improvement over unidirectional discriminator.}  
\label{tab:generation_quality}
\end{table*}

\subsection{Computational Gain}

\begin{figure}
\includegraphics[width=\columnwidth]{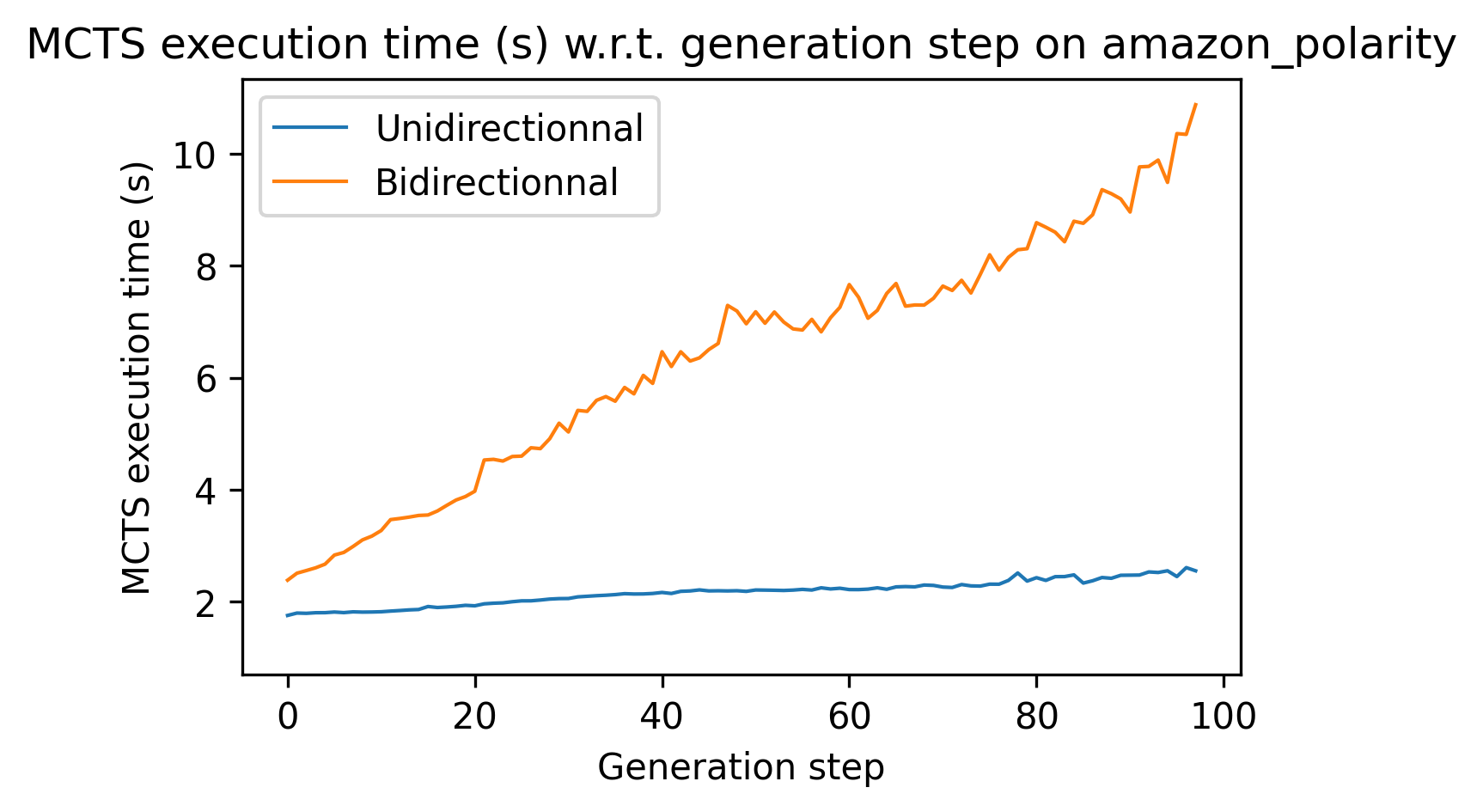}
\caption{Execution time of MCTS iterations (s) w.r.t. generation step (averaged over 10 batchs of 30 sequences)}
\label{fig:execution_time}
\end{figure}

Beyond generation accuracy, we are interested in computation complexity of the various models to be used in cooperative generation. %MCTS decoding. 
%The influence on the final sample of using one model over the other is of course critical in the choice of the discriminator to use. However, the cost is also really important, especially when considering a decoding strategy, where common methods such as greedy search, beam search and sampling are really fast. An high generation throughput is essentiel for the adoption of a decoding strategy. We thus study the difference in generation time of using a bidirectional and an unidirectional model when decoding using MCTS.
% In simple strategies, the difference in complexity of each model is directly measurable, as detailed for DAS in Section~\ref{sec:background}. When the strategy is more advanced, e.g using MCTS, the effective gain is harder to quantify. We thus study the difference in generation time of using a bidirectional and an unidirectional model when decoding using MCTS.
Fig.~\ref{fig:execution_time} reports MCTS execution times w.r.t. each generation step $t$ (i.e., time required to decode token at step $t$ of any sequence), using a bidirectional model compared to a unidirectional one. 
Unsurprisingly, since  the complexity is quadratic in the bidirectional case and only linear in the unidirectional one, the difference in generation time is significant, and increases linearly  w.r.t. the sequence length. % as shown in Figure~\ref{fig:execution_time}.
%The time needed to produce one token explodes as the sequence to complete becomes larger for %when using 
%the bidirectional setting,  %models
%while remaining %staying
%relatively constant with unidirectional ones.
%Note that the difference depends linearly on the number of MCTS iterations done per token.
Note also that this difference increases with the number of MCTS iterations. At last, we note that the number of MTCS iterations with unidirectional discriminator can be much more than doubled compared to the case of  bidirectional one,   while keeping the computational cost significantly lower, even for small text sequences.   

%For generative discriminators, there is two distinct cases when a node is selected during MCTS iteration. If one of its siblings has already been selected, we already have its discrimination score. If not, we need to compute it, which also gives its sibling scores which may be used in future iterations. Because the generation is done in batch, it is not suitable to not compute anything for certain element in the batch, even if the score needed is already available. For this reason, we decided that in such case, the available score is used and the compute is used to get the selected node children scores. This mean that the gain in compute is spent to explore more deeply, resulting in an execution time equal to that of a unidirectional discriminator. 
In the case of the generative discriminator, a great potential computational gain may arise from the fact that discrimination scores can be computed for every child of an expanded node at once.  %not
More specifically, while computing scores for each of the $|\mathcal{V}|$ children nodes would cost $|\mathcal{V}|$ forward passes in the case of discriminative classifiers, it only requires $|{\mathcal C}|$ forward passes for generative classifiers (i.e., one pass per class for getting all scores, rather than one pass per child node). Since usually $|{\mathcal C}|<<|\mathcal{V}|$, the use of generative discriminator could be way advantageous and allow to increase the number of MCTS iterations to expect to, at least,  fill the gap with accuracy results of discriminative approaches.    
%This gain is however easily measurable by counting the number of children of each node. When the first child of a node is selected, getting its score cost one forward pass with a discriminative model and $|\mathcal{C}|$ for the generative one. Then, when a sibling is selected, the cost is again one forward pass for the discriminative model but zero for the generative case. Thus, if we denote by $x_i$ the number of child of the node $i$, the amount of forward pass needed using a generative discriminator rather than a standard discriminator is defined by $\sum_{i, x_i > 0} x_i -  |\mathcal{C}|$. This implies that, the average number of children should be higher than $|\mathcal{C}|$ for generative discriminators to be beneficial. 

However, this potential gain heavily depends on the exploration of the tree and therefore the parameter $c_{puct}$. %, since scores can only be computed on the fly for discriminator approaches. 
If less than $|{\mathcal C}|$ children are considered at each level of the tree, then the generative approach is at least as costly as the discriminative one and can even be more costly. Indeed, % is considered  since more exploration implies more children. With 
we empirically observed that for usual value  $c_{puct}=3$, generative discriminators needs in average 1\,685 more forward passes on amazon\_polarity (where $|\mathcal{C}|$ is only 2), meaning there is more depth than width explorations. Increasing $c_{puct}$ decreases this difference but also the resulting generation accuracy. At $c_{puct}=15$, the accuracy already drops for 10 points and the difference is still to the disadvantage of GeDi for more than 600 forward pass. %Note that this difference also linearly depends on the number of MCTS iterations. 
These results show that generative discriminators are only beneficial if exploration is wider than deeper, which is not the case for MCTS operating points. This is consistent with GeDi results~\cite{DBLP:conf/emnlp/KrauseGMKJSR21}, which observed an important gain in a beam search decoding approach where the width is crucial. These new results suggest %Thus, it is necessary 
to seek at ways for  better leveraging this  GeDi potential with more  efficient  exploration in width of the MCTS or to use methods that do it by construction as beam search.%,  such as DAS for which this type of discriminator to be beneficial.

%\subsection{Variable length training}
%To ensure that discriminators are able to score unfinished sequences, a common practice is to train on sequences of random length. While it seems pretty logical, to the best of our knowledge, there is no experiments that support this. 
%Truncating training examples could be helpful to score the beginning of sequences but it might also be useless or even downright damaging by limitating the information during training. Studying the effective impact of such training allows to once and for all define if this specific training should be done or not.

%étudier si entraîner sur des entrées de tailles variables aide vraiment à classifier les débuts de texte

%%%%%%%%%%%%%%%%%%%%%%%%%%%%%%%%%%%%%%%%%%%%%%%%%%%%%%%%%%%%%%%
\section{Conclusion}

Cooperative generation has proven to be an effective way to augment traditional text generation with external information from a discriminator. While transformers with bidirectional attention are usually preferred for discriminative tasks, they are not auto-regressive and are therefore much more expensive when used to guide generation. 
Although a little less precise, unidirectional transformers allow to achieve very similar results for a much more reasonable and consistent cost. As a consequence, our study shows that unidirectional discriminators should be preferred for cooperative generation, for which slight accuracy drops can be % more than balanced by the use of greater numbers of MTCS iterations. %  in cases where bidirectionality is not needed to achieve good accuracy. 
balanced by reinvesting part of the computational gain.
Given the size of usual vocabularies, generative discriminators seem very interesting at first glance to allow wider search. However, while achieving similar results in terms of classification accuracy, %it seems that 
scoring the whole vocabulary comes at the price of a less informative signal. Moreover, although counter-intuitive, 
this width is not necessarily useful as shown by the search performed by the state-of-the-art Monte Carlo Tree Search, which usually explores more in depth than in width. Thus, such models will prove useful when used with methods that make particular use of this width information. We leave such explorations for future work. %We leave such MCTS modifications for future work. 

To allow reproduction and further experiments on this subject, the code used for our experiments is made available for the community at \href{https://github.com/NohTow/PPL-MCTS/tree/main/teammates}{https://github.com/NohTow/PPL-MCTS/tree/main/teammates}. % at https://anonymised-link

\bibliography{biblio}
\bibliographystyle{ACM-Reference-Format}

%%%%%%%%%%%%%%%%%%%%%%%%%%%%%%%%%%%%%%%%%%%%%%%%%%%%%%%%%%%%%%%%%%%%%%%%%%%%%%%
%%%%%%%%%%%%%%%%%%%%%%%%%%%%%%%%%%%%%%%%%%%%%%%%%%%%%%%%%%%%%%%%%%%%%%%%%%%%%%%
% APPENDIX
%%%%%%%%%%%%%%%%%%%%%%%%%%%%%%%%%%%%%%%%%%%%%%%%%%%%%%%%%%%%%%%%%%%%%%%%%%%%%%%
%%%%%%%%%%%%%%%%%%%%%%%%%%%%%%%%%%%%%%%%%%%%%%%%%%%%%%%%%%%%%%%%%%%%%%%%%%%%%%%
% \newpage
% \appendix
% \onecolumn
% \section{Appendix}

%%%%%%%%%%%%%%%%%%%%%%%%%%%%%%%%%%%%%%%%%%%%%%%%%%%%%%%%%%%%%%%%%%%%%%%%%%%%%%%
%%%%%%%%%%%%%%%%%%%%%%%%%%%%%%%%%%%%%%%%%%%%%%%%%%%%%%%%%%%%%%%%%%%%%%%%%%%%%%%

\end{document}